# A PRISMA Driven Systematic Review of Publicly Available Datasets for Benchmark and Model Developments for Industrial Defect Detection

Can Akbas[1], Irem Su Arin[1], and Sinan Onal[1*]
[1]Department of Industrial Engineering, Southern Illinois University, Edwardsville, USA

*Corresponding author: Sinan Onal, sonal@siue.edu

**Abstract:** Recent advancements in quality control across various industries have increasingly utilized the integration of video cameras and image processing for effective defect detection. A critical barrier to progress is the scarcity of comprehensive datasets featuring annotated defects, which are essential for developing and refining automated defect detection models. This systematic review, spanning from 2015 to 2023, identifies 15 publicly available datasets and critically examines them to assess their effectiveness and applicability for benchmarking and model development. Our findings reveal a diverse landscape of datasets, such as NEU-CLS, NEU-DET, DAGM, KolektorSDD, PCB Defect Dataset, and the Hollow Cylindrical Defect Detection Dataset, each with unique strengths and limitations in terms of image quality, defect type representation, and real-world applicability. The goal of this systematic review is to consolidate these datasets in a single location, providing researchers who seek such publicly available resources with a comprehensive reference.

## 1. Introduction

In In the field of industrial production and quality management, the critical role of defect detection in ensuring product integrity and adherence to quality standards is widely recognized. This process is crucial for maintaining customer satisfaction, ensuring cost efficiency, and upholding safety and brand reputation. Traditionally, defect detection has relied on manual inspection, which, despite its merits, is fraught with limitations such as vulnerability to human error, inconsistency, time consumption, and limited capability in detecting micro-level or complex defects. These limitations, particularly pronounced in high-volume production environments, necessitate the exploration of more advanced, automated solutions to enhance accuracy, efficiency, and consistency in defect detection.

The advent of image processing technology has marked a significant advancement in quality control processes. With a variety of image processing models, this technology offers enhanced capabilities in improving image clarity and facilitating detailed quality assessments. However, the diverse nature of manufacturing processes demands that these models be tailored to specific production types, a requirement that introduces the challenge of model variability across different datasets. This underscores the need for a comprehensive evaluation of model efficacy.

Recognizing the importance of standardized datasets in overcoming these challenges, there has been an increasing emphasis on developing unified, benchmark datasets for the fair and effective assessment of image processing models. These datasets enable comparative analysis of model performance and help elucidate the relative strengths and weaknesses of different approaches. Yet, there exists a significant gap in the literature, particularly in terms of systematic, up-to-date reviews of these benchmark databases.

This systematic review aims to bridge this gap by assembling and evaluating a comprehensive list of widely used and reputable public datasets in the field of industrial defect detection. Our goal is to offer a consolidated resource that supports researchers and industry practitioners in their experimental and operational endeavors. Conducted in accordance with the PRISMA 2020 guidelines (Page et al., 2021), this review not only seeks to synthesize the current state of benchmark and standardized datasets for industrial defect detection but also to provide a comprehensive evaluation of the databases and datasets used in this context. We aim to understand their characteristics, strengths, and limitations, and how these datasets have been employed in recent research to enhance defect detection algorithms and methodologies.

By systematically analyzing these resources, this review will contribute significantly to the field of industrial defect detection. We aim to offer insights into the effectiveness of various datasets, serving as a guide for selecting and utilizing the most appropriate resources for specific defect detection needs in industrial manufacturing, thereby enhancing the reliability and efficiency of quality control processes.

## 2. Materials and Methods
### 2.1. The PRISMA Protocol

In this systematic review, we thoroughly adhered to the Preferred Reporting Items for Systematic Reviews and Meta-Analyses (PRISMA) 2020 guidelines, ensuring a comprehensive, transparent, and unbiased approach. Our methodical process was divided into four critical phases: identification, screening, eligibility, and inclusion, in line with PRISMA's structured methodology. Initially, an extensive search across various databases identified a broad range of potentially relevant studies. This was followed by a preliminary screening of titles and abstracts to filter out unrelated studies. We then conducted a detailed review of the full texts of the remaining studies, applying stringent inclusion and exclusion criteria to select the most relevant and high-quality studies for our review. The final phase involved a thorough analysis and inclusion of studies that met all our criteria, ensuring a robust and valid compilation of data for synthesis. This rigorous adherence to the PRISMA guidelines has been instrumental in maintaining the integrity and utility of our findings in the realm of industrial defect detection.

### 2.2. Eligibility and Selection Process

This systematic review strictly adhered to the PRISMA guidelines, beginning with the meticulous definition of inclusion and exclusion criteria. Studies were included if they focused on image-based industrial defect detection, published in English between 2015 and 2023 due to the focus on capturing the most recent advancements and methodologies in the field of industrial defect detection. This time frame was selected to ensure that the review encompassed contemporary datasets and technologies, reflecting the current state of the art and recent innovations. It allowed for the inclusion of studies that incorporated the latest machine learning algorithms, image processing techniques, and defect detection models, which are crucial for understanding the present capabilities and future directions of this rapidly evolving field. Studies were excluded if unrelated to image processing or lacking empirical data because the primary objective of this review was to assess the efficacy and application of image processing techniques in industrial defect detection. Focusing on image processing ensured that the review remained concentrated on the most relevant and technologically advanced approaches in the field. Moreover, the requirement for empirical data was crucial for grounding our analysis in concrete, verifiable results, allowing for a more robust and credible evaluation of the datasets and methodologies. Studies without empirical data or not centered on image processing would have diverged from the core purpose of our review, potentially diluting the relevance and specificity of our findings in the context of industrial defect detection.

The selection process involved two independent reviewers conducting a thorough screening of titles and abstracts, followed by a full-text review to finalize the study selection. Discrepancies were resolved through consensus or by consulting a third reviewer, ensuring an unbiased and comprehensive inclusion of relevant studies.

### 2.3. Search Strategy and Data Collection

An exhaustive literature search was performed across key databases like ACM Digital Library, Scopus, ProQuest, and IEEE Xplore due to their extensive coverage of literature in technology and engineering. The search, last updated on [insert date], utilized a combination of keywords and Boolean operators, with filters to exclude pre-2015 publications. The search strategies for each database are detailed in the supplementary material. Data collection was systematically carried out by two reviewers, focusing on extracting vital information such as study characteristics, methods, outcomes, and findings. To refine our search, we added terms specifically related to defect detection in industrial contexts, including "image data set" OR "image dataset" OR "visual dataset" OR "image database") AND ("defect detection" OR "defect inspection" OR "surface defect detection" OR "defect classification" OR "scratch detection") AND ("inspect" OR "product" OR "production" OR "manufacturing" OR "assembly"). In cases of missing or unclear data, study authors were contacted for additional information or clarification.

### 2.4. Synthesis and Bias Assessment

Studies were grouped for the syntheses based on several key criteria to ensure a coherent and meaningful analysis. First, they were categorized by the industrial sector they focused on, such as automotive, electronics, or textiles, allowing us to assess and compare the specific requirements and challenges within each domain. Secondly, the studies were grouped according to the type of defects they addressed — for example, surface defects, structural defects, or specific anomalies like cracks and scratches. This categorization facilitated a targeted analysis of defect detection effectiveness across different defect types. Additionally, we grouped studies based on the methodologies employed, such as traditional image processing techniques versus advanced machine learning algorithms, to evaluate and contrast their relative efficacies. Finally, the datasets used in each study were another criterion for grouping, distinguishing between synthetically generated datasets and those derived from real-world industrial environments. This approach allowed us to synthesize the data in a way that highlighted not only the diversity of the research in this field but also the common trends and emerging patterns, providing comprehensive insights into the state of industrial defect detection.

To assess the risk of bias in the included studies, a rigorous evaluation was conducted using the Cochrane Risk of Bias Tool as the assessment tool. Each study was carefully scrutinized for factors such as the methodology used, the selection of datasets, the objectivity of the analysis, and the transparency of reporting results. This assessment was carried out independently by two reviewers to minimize subjective bias and ensure a comprehensive evaluation. In cases of disagreement, a consensus was reached through discussion or consultation with a third reviewer. The findings from this risk of bias assessment provided crucial insights into the reliability and validity of the study results, informing the overall synthesis of the review. It also helped in identifying potential areas of methodological improvement for future research in the field of industrial defect detection, ensuring that subsequent analyses and developments can be based on robust, unbiased foundations.

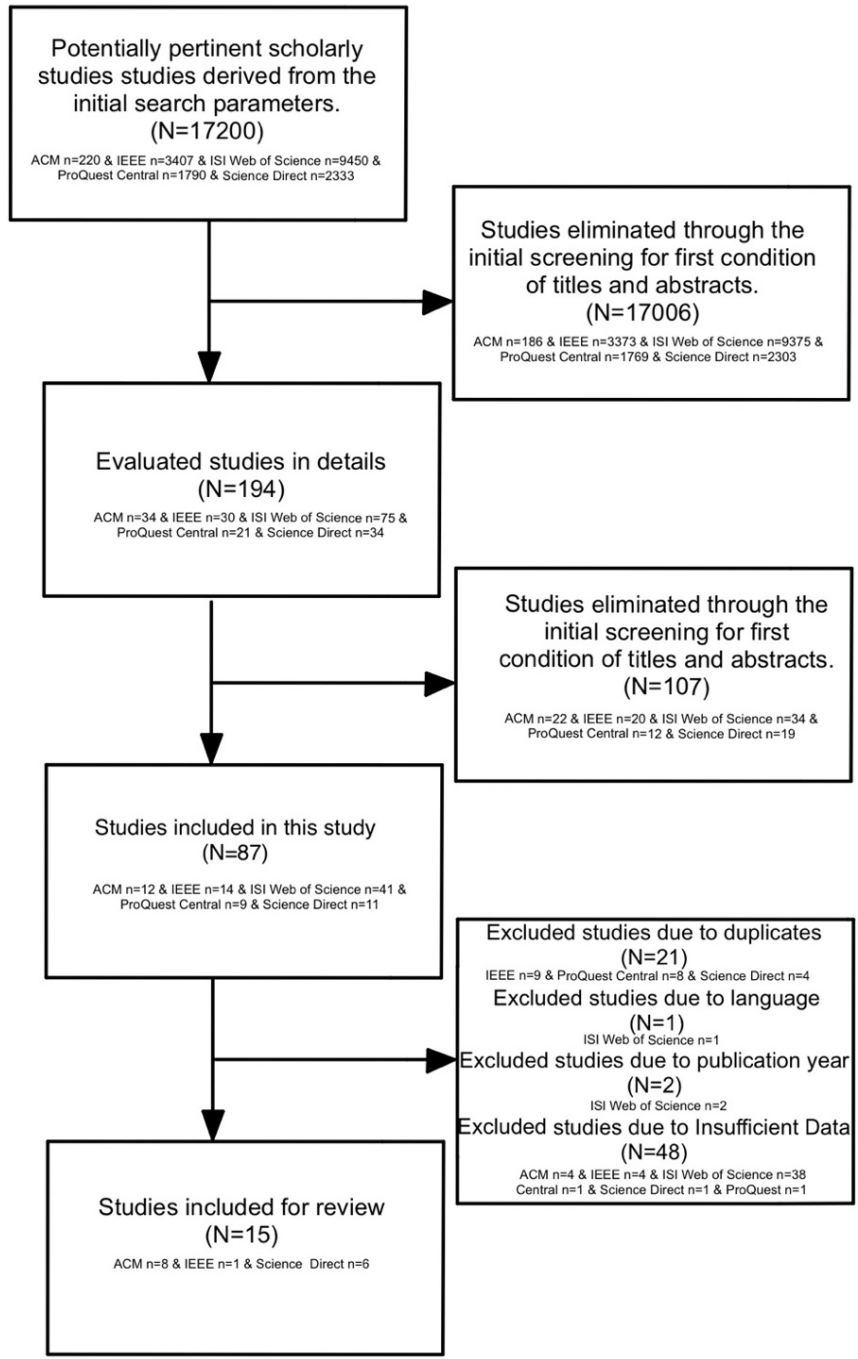

Figure 1: PRISMA systematic flowchart

Table 1: Summary of included studies (n=15)

| Item | Studies | Dataset | # of Images | Image size | Industry | Defect Type |
|---|---|---|---|---|---|---|
| 1 | (Cao et al., 2023) | NEU-CLS | 1,800 | 200x200 | Steel | Defect Classification |
| 2 | (Yu et al., 2023) | NEU-DET | 1,800 | 200x200 | Steel | Surface Defect |
| 3 | (Qin et al., 2020) | DAGM | 16,100 | 512x512 | Textured Surface | Surface Defect |
| 4 | (S. Xu & Shao, 2023)6/11/2024 3:11:00 PM | NEU-DET | 1,800 | 200x200 | Steel | Surface Defect |
| 5 | (Liu et al., 2022) | NEU-DET | 1,800 | 200x200 | Steel | Surface Defect |
| 6 | (Z. He & Q. Liu, 2020) | DAGM | 16,100 | 512x512 | Textured Surface | Surface Defect |
| 7 | (Z. Lin et al., 2019) | 6/11/2024 3:11:00 | 1,800 | 200x200 | Steel | Surface Defect |
|   |   | PMNEU-DET | 16,100 | 512x512 | Textured Surface | Surface Defect |
|   |   | DAGM |   |   |   |   |
| 8 | (Sharma & Kuppili, 2023) | KolektorSDD | 399 | 500x1240 500x1270 | Steel | Surface Defect |
| 9 | (Singh et al., 2023a) | Severstal | 18,100 | 1600x256 | Steel | Surface Defect |
| 10 | (Yu et al., 2023) | Tianchi Fabric | 8,000 | 1037x412 | Textile | Surface Defect |
|   |   | Pekin University PCB | 1,386 | 3034x1586 | PCB | Surface Defect |
| 11 | (Shafi et al., 2022) | Hollow Cylindrical | 2,142 | 640x480 | Cylindrical Surface | Surface Defect |
| 12 | (C. Xu et al., 2023) | COCO | 330,000 | 640x480 | Computer Vision | - |
| 13 | (Niu et al., 2023) | COCO | 330,000 | 640x480 | Computer Vision | - |
| 14 | (Liyun et al., 2020) | Pascal VOC | 58,716 | 500x375 | Computer Vision | - |
| 15 | (Zhong et al., 2023) | ModelNet40 | 12,311 | 3D | Computer Vision | - |

## 3. Results

In this systematic review, conducted using the PRISMA methodology, we meticulously assessed the efficacy and characteristics of datasets used in industrial defect detection. Our initial analysis focused on the applicability of these datasets in defect detection algorithms, taking into account factors like dataset size, diversity of defects, and real-world relevance. The datasets, categorized by application domains such as steel, electronics, and textiles, revealed distinct requirements and challenges in each domain, with a notable emphasis on surface defect detection in the automotive sector due to the critical nature of these defects. Our evaluation also noted an evolution towards more complex and real-world scenarios in newer datasets, presenting challenges to the adaptability and robustness of existing algorithms.

The extensive search for relevant literature resulted in 17,200 articles, which were then narrowed down through a rigorous screening process. This process led to the exclusion of the majority of these articles for reasons such as irrelevance, duplication, language, and publication date, ultimately leaving 15 articles that met our strict inclusion criteria. These articles, critically reviewed for their methodologies, types of datasets used, and defect detection outcomes, form the foundation of our review. Our analysis identified a need for more diversified datasets encompassing a broader range of defects and manufacturing processes and highlighted a gap in datasets with annotated defect information, critical for training and testing advanced machine learning algorithms. The PRISMA flowchart (Figure 1) provides a visual overview of the methodological rigor and compliance with the PRISMA guidelines.

This review has not only provided a comprehensive overview of the current state of datasets in industrial defect detection, revealing their strengths and limitations, but also offered insights for future improvements in the field. We observed that a significant number of datasets, often used in industrial defect detection and computer vision, were proprietary and not publicly available. Our focus was on datasets that were openly shared and recognized as benchmarks in the field, leading to a detailed compilation of these key resources, as summarized in Table 1. The selected articles, encompassing various aspects like surface defects in the steel industry, reflect the diverse applications and methodologies employed in current research.

## 4. Discussions

One notable dataset, emerging prominently from our review, is from Northeastern University (NEU) in China (Bao et al., 2021). This dataset has established itself as a widely utilized resource for surface defect detection and classification, particularly in the context of hot-rolled steel strips. It comprises 1,800 grayscale images, meticulously capturing six common surface defects, thereby ensuring a balanced representation of each defect type. The comprehensiveness and variety of defects in the NEU dataset make it an invaluable resource for developing and testing advanced defect detection algorithms.

The versatility and applicability of the NEU dataset have been showcased in various studies, highlighting its adaptability across different research contexts and methodologies. For instance, (Yu et al., 2023) utilized this dataset in complex scenario testing, demonstrating its capability in simulating real-world manufacturing conditions. Similarly, (Z. Lin et al., 2019) and (S. Xu & Shao, 2023) leveraged the NEU dataset for migration learning and few-shot defect detection, underscoring its effectiveness in training machine learning models that require less data to learn from. This adaptability is crucial, given the evolving nature of manufacturing processes and the increasing complexity of defect detection tasks. Furthermore, the employment of the NEU dataset by (Liu et al., 2022) in their research provides insights into its role in enhancing algorithmic precision. This is especially significant in industries where even minor surface defects can lead to significant repercussions, emphasizing the need for highly accurate defect detection systems. The NEU dataset's diverse applications in various studies not only attest to its robustness and reliability but also highlight its contribution to advancing the field of defect detection in industrial settings.

The widespread use and the results derived from these studies suggest that the NEU dataset can serve as a benchmark for future dataset developments. It sets a precedent for what comprehensive and application-specific datasets should encompass, particularly in the realm of defect detection in manufacturing. Consequently, this review underscores the NEU dataset's instrumental role in driving forward the research and development of more nuanced and sophisticated defect detection methodologies that can cater to the ever-growing demands of modern manufacturing industries. The NEU dataset's contributions to the field extend beyond its immediate applications. It provides a foundation for future research, encouraging the development of more diverse and comprehensive datasets that can address a broader spectrum of manufacturing defects. Such advancements are essential for keeping pace with the technological advancements and increasing complexity in industrial manufacturing processes.

The German Association for Pattern Recognition, commonly referred to as DAGM (Weimer et al., 2016), has made a significant contribution to the field of industrial defect detection by providing an invaluable dataset. This dataset is particularly notable for being artificially generated, a design choice that allows it to effectively mimic real-world problems. Such a synthetic approach is crucial in creating controlled environments where specific variables can be manipulated to assess the robustness and efficacy of classification algorithms, especially in industrial optical inspection scenarios.

The DAGM dataset's impact on advancing classification algorithms is evident from its widespread utilization in research. For example, (Z. He & Q. Liu, 2020) leveraged this dataset to evaluate deep regression neural networks, demonstrating its applicability in sophisticated, AI-driven inspection systems. Similarly, (Qin et al., 2020) and (Z. Lin et al., 2019) employed the DAGM dataset to test lightweight surface defect detection systems. These applications underline the dataset's versatility in catering to both computationally intensive models and more streamlined, efficient algorithms. The artificial nature of the DAGM dataset offers several advantages. It provides a high degree of control over the defect types and their characteristics, enabling researchers to systematically evaluate and refine their algorithms. This is particularly beneficial in the early stages of algorithm development, where controlled conditions are necessary to understand the fundamental strengths and weaknesses of new methods. Furthermore, the DAGM dataset acts as a critical benchmark against which the performance of emerging technologies can be measured. It sets a standard for the kind of precision and adaptability required in future datasets and defect detection algorithms. Additionally, the use of the DAGM dataset in various research contexts has highlighted its potential as a training tool for machine learning models. By encompassing a wide range of simulated defects, it helps in training algorithms to be more adaptive and responsive to a variety of defect types, thereby enhancing their applicability in real-world manufacturing settings. The DAGM dataset not only serves as a testbed for emerging technologies but also provides a blueprint for the development of future datasets that can more accurately represent the complex nature of industrial defects. Its role in the advancement of classification algorithms for industrial optical inspection cannot be overstated, as it continues to push the boundaries of what is possible in automated defect detection systems.

The Kolektor Surface Defect Dataset (KolektorSDD), furnished by the Kolektor Group (Tabernik et al., 2020), represents a significant advancement in the realm of industrial defect detection. This dataset stands out due to its collection of images featuring defective production items, all captured within a controlled industrial environment. The real-world nature of these images, combined with the controlled conditions of their capture, provides an optimal balance for testing and developing defect detection algorithms.

KolektorSDD's utility is exemplified in its application across various research studies, particularly those focusing on texture defect detection in industrial products. Researchers like (Sharma & Kuppili, 2023) have utilized this dataset to test and validate novel approaches in this area. The dataset's real-world images allow for a realistic assessment of the performance of these novel algorithms, offering valuable insights into their practical applicability in actual manufacturing settings. One of the key strengths of the KolektorSDD lies in its specificity and focus. By concentrating on a specific type of defect — texture defects in this case — the dataset allows for a more in-depth analysis and understanding of these specific challenges. This is crucial in industries where even minor surface irregularities can lead to significant product quality issues. The high-quality and high-resolution images in the KolektorSDD provide a detailed view of such defects, making it an excellent resource for developing more precise and accurate defect detection methods. Moreover, the KolektorSDD's contribution extends beyond mere defect identification.

It facilitates the development of algorithms that can distinguish between different types of surface imperfections, a capability that is increasingly important as manufacturing processes become more complex and nuanced. The dataset's use in developing and testing these advanced algorithms underscores its value as a tool for enhancing the overall quality control process in industrial manufacturing. In essence, KolektorSDD, not only serves as a crucial resource for current research but also sets a benchmark for the quality and type of data that should be made available for future developments in the field. It exemplifies the importance of having detailed, real-world datasets that are closely aligned with the practical needs of the manufacturing industry, thereby playing a pivotal role in driving forward technological advancements in industrial defect detection.

Beyond the primary databases previously discussed, our systematic review uncovered several other datasets that, while mentioned in the literature, have varied applications and suitability for industrial defect detection. The Severstal Steel Surface Defect Detection Dataset (Alexey Grishin, 2019), hosted on Kaggle, is a notable example. Although it offers a comprehensive collection of steel surface defects, it was not utilized in certain studies due to specific research requirements or the nature of the defects represented. This highlights the importance of dataset selection in alignment with the specific objectives and requirements of each study.

The PCB Defect Dataset from Peking University (Niu et al., 2023), focusing on printed circuit board defects, addresses the intricate requirements of electronic manufacturing, where even minuscule defects can have significant consequences. This dataset's specificity for PCBs highlights the trend towards more specialized datasets in the field. Additionally, the Hollow Cylindrical Defect Detection Dataset (Shafi et al., 2022), developed to bridge the gap in cylindrical surface defect detection, exemplifies the ongoing efforts to create datasets that address less common but equally important types of industrial defects.

Additionally, datasets from 12 to 15 in the table, such as the Common Objects in Context database (COCO) (T.-Y. Lin et al., 2014), Pascal VOC (Hoiem et al., 2009), and ModelNet40 (Sun et al., n.d.), although extensively used in computer vision, were not directly related to industrial defect detection. Their inclusion in some of the reviewed studies for model training, however, highlights the interdisciplinary nature of research methodologies in this field. For example, the COCO database, renowned in the field of computer vision. However, its general-purpose nature and focus on everyday objects have rendered it less suitable for specialized industrial applications, underlining the need for more industry-specific datasets in defect detection research. Similarly, the PASCAL VOC dataset, while invaluable in object classification tasks, was found to be inapplicable for industrial defect detection due to its general content and lack of industrial specificity.

These additional datasets, while not always directly applicable to every industrial defect detection scenario, contribute to the broader understanding of the varied requirements and challenges in this field. They collectively emphasize the need for diverse, targeted datasets that can support the development of advanced defect detection systems tailored to specific industrial applications. This diversity in datasets not only enriches the field but also guides future research directions, encouraging the development of more nuanced and application-specific defect detection solutions.

The findings of this review have several implications. For practitioners in industrial defect detection, the reviewed datasets provide a benchmark for developing and testing new algorithms. For policymakers, understanding the limitations and strengths of these datasets can guide decisions on funding and support for research in developing more comprehensive and diverse datasets. Future research should focus on expanding the range and diversity of defect detection datasets, including a broader array of materials and defect types. There is also a need for datasets with annotated defect information to facilitate the training of more advanced machine learning algorithms. Additionally, exploration into the integration of datasets from various domains, as seen with the inclusion of general computer vision datasets in some studies, may open new avenues for innovative defect detection methodologies.

The present scoping review is not without limitations. First, our review focused primarily on articles available in English, potentially overlooking valuable insights from studies published in other languages. This language restriction may limit the comprehensiveness of our analysis. Second, the inclusion criteria, while rigorous, may have inadvertently excluded relevant studies that did not meet the specific terms of our search strategy or were published outside the databases we surveyed. Additionally, our reliance on publicly available datasets means that proprietary or unpublished datasets, which could contain important information, were not considered. This could lead to a certain degree of selection bias. Furthermore, the rapid evolution of technology in industrial defect detection means that our review might not capture the very latest advancements in the field. Finally, while we endeavored to critically appraise the quality of the included studies, inherent biases within these studies may still influence our findings. Despite these limitations, we believe our review provides valuable insights and a comprehensive overview of the current state of datasets in industrial defect detection, serving as a foundation for future research in this area.

## 5. Conclusion

Our systematic review, conducted in strict adherence to the PRISMA guidelines, delivers a comprehensive analysis of the current landscape of publicly available datasets in industrial defect detection. This careful evaluation encompassed 15 studies, each offering critical insights into the characteristics and efficacy of a variety of datasets. We discovered a diverse array of datasets, each with its own set of strengths and weaknesses. Key datasets such as NEU-CLS, NEU-DET, DAGM, KolektorSDD, PCB Defect Dataset, and the Hollow Cylindrical Defect Detection Dataset have significantly contributed to advancements in the field. Their primary focus has been on surface defect detection in materials like steel and textured surfaces. The deployment of these datasets in various research initiatives underscores the critical role of high-quality, high-resolution images in the development and efficacy assessment of defect detection algorithms.

Our analysis also highlights the necessity for the continuous development and updating of datasets. Keeping abreast of the ever-evolving manufacturing processes and technologies in defect detection is vital. Integrating datasets from diverse industrial sectors and defect types is essential to create more robust and adaptable defect detection systems. These benchmark datasets are invaluable in providing a common ground for comparing proposed models, thus driving advancements in defect detection. However, as indicated by (Singh et al., 2023b). in their study 'Vision-based system for automated image dataset labelling and dimension measurements on shop floor,' while publicly available datasets are commonly used for training computer vision models, case-specific data often yield superior results. Nonetheless, benchmarks and standard datasets remain fundamental for the evaluation and testing of new industrial defect detection models.

**Conflict-of-Interest Statement**

The authors declare that there are no conflicts of interest regarding the publication of this paper. This assertion includes, but is not limited to, financial, personal, or professional relationships that could be construed as potential conflicts of interest.